\documentclass{article}

\usepackage[final]{neurips_2023}

\usepackage[utf8]{inputenc} 
\usepackage[T1]{fontenc}    
\usepackage{hyperref}       
\usepackage{url}            
\usepackage{booktabs}       
\usepackage{amsmath}
\usepackage{amsfonts}       
\usepackage{nicefrac}       
\usepackage{microtype}      
\usepackage{xcolor}         
\usepackage{graphicx}
\usepackage{natbib}
\usepackage{doi}
\usepackage{svg}
\usepackage{appendix}

\title{Balancing utility and cognitive cost in social representation}

\author{Max Taylor-Davies \\
	School of Informatics \\
	University of Edinburgh \\
	\texttt{s2227283@ed.ac.uk} \\
    \And
	Christopher G. Lucas \\
	School of Informatics \\
	University of Edinburgh \\
}

\begin{document}

\maketitle

\begin{abstract}
To successfully navigate its environment, an agent must construct and maintain representations of the other agents that it encounters. Such representations are useful for many tasks, but they are not without cost. As a result, agents must make decisions regarding how much information they choose to store about the agents in their environment. Using selective social learning as an example task, we motivate the problem of finding agent representations that optimally trade off between downstream utility and information cost, and illustrate two example approaches to resource-constrained social representation.  
\end{abstract}

\section*{Representing agents under cost constraints}

In order to produce adaptive behaviour, an agent must acquire and maintain an internal representation of its environment \citep{Craik1943, Tolman1948CognitiveMI, Wilson2014OrbitofrontalCA}. Furthermore, unless condemned to an entirely solitary existence, we can expect that many environments encountered by a hypothetical agent will contain \textit{other agents}. Much as our agent should represent the rest of the environment, we expect that it ought also to be able to represent these other agents. Humans do this---in fact, it seems we are intrinsically motivated to form mental representations of the other people we encounter \citep{Dennett1987TheIS, MalleFundamentalTools, baker2017rational}. We use these representations for a variety of different purposes: understanding the strengths and weaknesses of a colleague to effectively collaborate with them; determining whether a stranger should be treated as friend or foe; or predicting the plays of a chess opponent in order to beat them. In an ideal world, we would maintain representations that encode all possible available information about every agent in our environment. But real agents, whether biological or artificial, will inevitably have to contend with limits on their cognitive resources. Inspired by a growing body of work in computational cognitive science that models human cognition through the lens of resource rationality \citep{Lieder2019ResourcerationalAU, Bhui2021ResourcerationalDM}, we consider the problem of developing social representations that balance downstream utility against cognitive cost. 

\section*{Social decision tasks}
To consider the idea of using social representations for decision-making, we develop the general construct of a \textit{social decision task}---which describes any decision task where the optimal strategy depends on some representation of some set of other agents. That is, adapting the notation of Markov Decision Processes (MDPs) \citep{Sutton1998, Puterman1994}, the optimal policy (in terms of maximising expected return) is conditional on the value of some social representation $\chi$: $\pi^*(s,a) = \pi^*(s,a ; \chi)$. To consider the utility of a particular social representation $\chi$, first let $\Pi(\chi)$ be the class of possible policies whose output depend on the value of $\chi$. We can then consider the utility of $\chi$ as the expected return induced by the \textit{best} member of $\Pi(\chi)$:
\begin{equation}\label{eq:chi-utility}
    U(\chi) = \max_{\pi \in \Pi(\chi)}\mathbb{E}[R \ | a \sim \pi(s,a ; \chi)]
\end{equation}
Where $R$ is the total (possibly discounted) return achieved from executing actions $a$ in the environment. Equation \ref{eq:chi-utility} tells us, given access to social representation $\chi$, how well we expect to perform on a particular task assuming that we're able to make the best possible use of the information in $\chi$? Given some function $C : \chi \to \mathbb{R}$ that measures cognitive cost, we then define the cost-adjusted utility of $\chi$ as
\begin{equation}\label{eq:cost-adjusted-utility}
    U'(\chi) = (1 - \lambda)U(\chi) - \lambda C(\chi)
\end{equation}
where $\lambda \in [0,1]$ is a parameter that trades off between utility and cost. A representation $\chi^*$ is then considered optimal if it satisfies $\chi^* = \arg\max_\chi U'(\chi)$. This optimality criterion is similar to the objectives used within work on capacity-limited Bayesian decision-making and RL, such as \citet{Arumugam2023BayesianRL}. The key difference (beyond our explicit focus on social representations) is that we are interested not so much in the cognitive cost of converting representations into behaviour, but in the cost the representations themselves. In general, we expect this be a combination of the cost involved in \emph{acquiring} a representation (i.e. inferring it from observation), and the cost involved in \emph{storing} it---for now we simply group both under entropy (for more details see Appendix \ref{appx:entropy}), i.e. $C(\chi) = H[\chi]$. This makes intuitive sense as an objective---essentially saying that we want to maximise our expected task performance while minimising the amount of information we store about other agents. This is related to the objective used in \citet{Abel2019StateAA}, which considers a similar tradeoff for the problem of state abstraction in apprenticeship learning. 

\section*{Selective social learning as an example downstream task}\label{sec:selective-imitation}
As a simple example of a social decision task, we consider \textit{selective social learning}, a common component of human and animal social behaviour which involves the problem of identifying the `best' agent to learn from within a given environment \citep{Rendell2011, Heyes2016}. In our framing of this task, the learning agent $\alpha^\text{ego}$ is situated within some environment populated by a number of other agents, with the ability to observe their behaviour (in the form of state-action transitions). At each trial, $\alpha^\text{ego}$ selects an agent $\alpha^\text{target}$. All agents then execute some sequence of behaviour within the environment, visible to $\alpha^\text{ego}$. $\alpha^\text{ego}$ executes the same behaviour observed from the chosen $\alpha^\text{target}$, and receives some reward corresponding to their own internal utility function $\mathbf{u}^\text{ego}$. A general policy for selecting social learning targets on the basis of representations is given by
\begin{equation}\label{eq:weighted-agent-selection}
    \text{Pr}(\alpha^\text{target} = \alpha^{(m)}) \propto \exp\bigg( \frac{w(\alpha^{(m)}; \chi)}{\beta^\text{ego}} \bigg) 
\end{equation}
where $w$ is some function that weights agents based on the value of $\chi$ and $\beta^\text{ego}$ gives the \emph{decision noise} of the social learning agent.

\begin{figure}
    \centering
    \includegraphics[width=10cm]{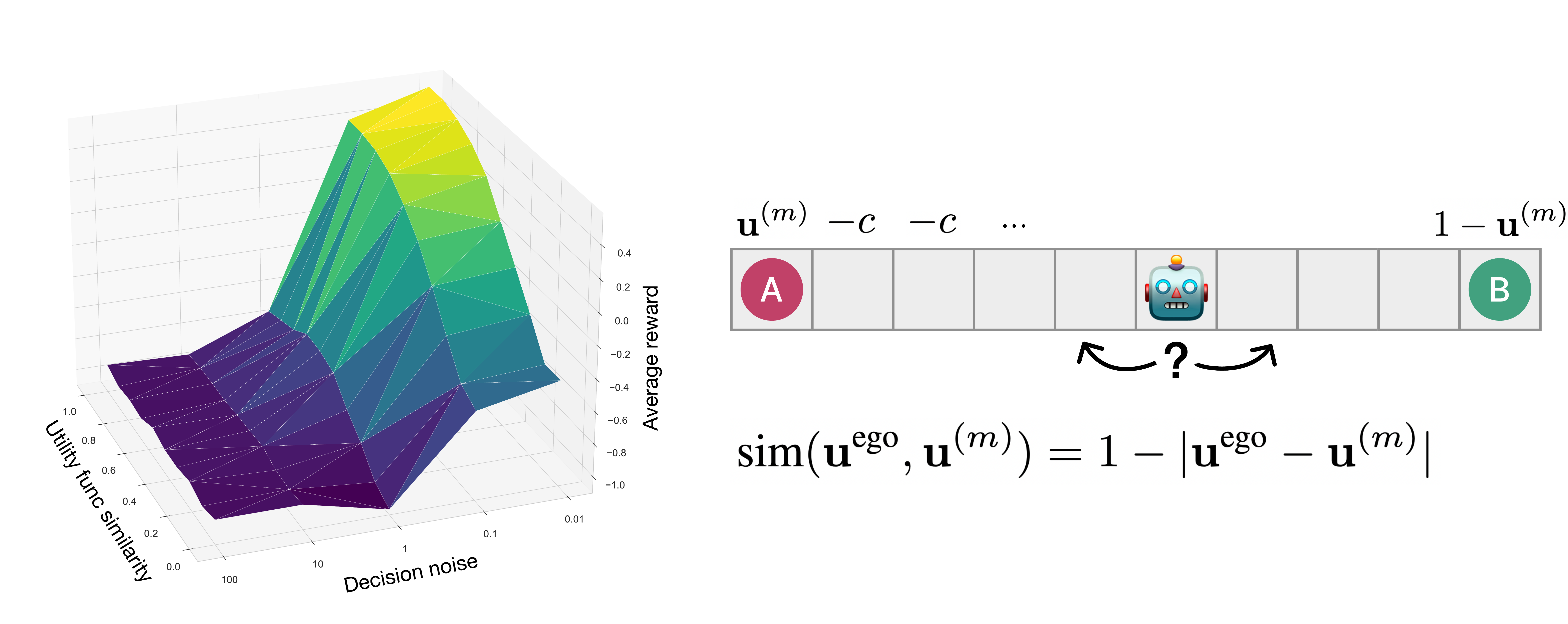}
    \caption{We generate several agents with randomly sampled scalar utility functions $\mathbf{u}^{(m)} \in [0,1]$, which control each agent's relative preference for the two states labelled A and B. From each agent we sample a number of trajectories using different values of the decision noise parameter $\beta^{(m)}$. We then plot social learning reward as a function of the \emph{similarity} between $\mathbf{u}^{(m)}$ and $\mathbf{u}^\text{ego}$, and $\beta^{(m)}$, averaged over both repeated trials and different values of $\mathbf{u}^\text{ego}$.}
    \label{fig:surfaces-minimal}
\end{figure}

We can use Equation \ref{eq:weighted-agent-selection} to express different selection strategies by changing the weighting function $w$. Of course, the optimal choice of $w$ will depend on the type of decision-making employed by the other agents. We assume that each observed agent attempts to maximise its own return in a noisily rational manner, according to a Boltzmann-style policy with decision noise $\beta^{(m)}$. Under this model, the social learner's expected reward should be optimised by selecting an agent such that we maximise some measure of the \emph{similarity} between $\mathbf{u}^\text{target}$ and $\mathbf{u}^\text{ego}$ while minimising $\beta^{(m)}$. We can demonstrate this through simulation using a simplified one-dimensional gridworld environment---Fig. \ref{fig:surfaces-minimal} shows that the social learner's return indeed increases with increasing utility function similarity and decreasing $\beta^{(m)}$. For the remainder of this paper, we will make the assumption that all observed agents are equivalently rational, and thus set aside the decision noise parameter. We therefore consider optimal social representations of the form $\chi^{(m)} = \hat{\mathbf{u}}^{(m)}$ where $\hat{\mathbf{u}}^{(m)}$ is some approximation of $\mathbf{u}^{(m)}$. 

To study strategies for selective social learning, we use a contextual bandit setting. At each trial, each observed agent $m$ is placed at a random tile within a 2D grid, chooses an destination tile to travel to, and then receives a reward $r = \mathbf{u}^{(m)}_\text{dest} - c\big(|x_\text{dest} - x_\text{start}| + |y_\text{dest} - y_\text{start}|\big)$ where $c \geq 0$ is a constant scalar that determines the cost of taking one step on the grid. Following the design of \citet{Wu2018GeneralizationGH} and \citet{witt2023social}, we generate spatially correlated 2D utility functions (i.e. nearby tiles generally yield similar rewards) by sampling from a Gaussian process prior with a radial basis function kernel. The social learner selects targets according to Eq. \ref{eq:weighted-agent-selection}, with weights given by 
\begin{equation}\label{similarity-weights}
w(\alpha^{(m)}; \chi) = w(\alpha^{(m)}; \hat{\mathbf{u}}^{(m)}) = \text{sim}(\mathbf{u}^\text{ego}, \hat{\mathbf{u}}^{(m)}) = \frac{\mathbf{u}^\text{ego} \cdot \hat{\mathbf{u}}^{(m)}}{|\hat{\mathbf{u}}^{(m)}||\hat{\mathbf{u}}^{(m)}|}
\end{equation}

\section*{Experiment 1: state aggregation}
\begin{figure}
    \centering
    \includegraphics[width=12.5cm]{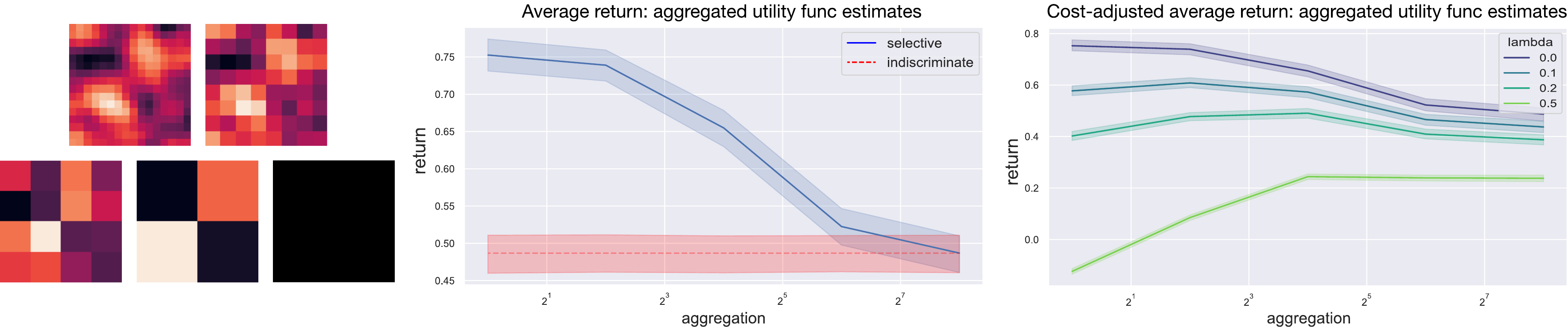}
    \caption{Left: illustration of state-aggregated estimates of an example 2D utility function (lighter grid squares indicate higher values). Middle: average return from selective social learning using state-aggregated value function estimates as agent representation (with indiscriminate selection as a baseline). Right: cost-adjusted return for aggregated representations at different values of $\lambda$. Since we are interested only in the tradeoff between return and cost (entropy), rather than each quantity's absolute values, both are individually normalised to lie in [0,1].}
    \label{fig:2d-bandit-results}
\end{figure}

If $\alpha^\text{ego}$ has to choose between only a small number of agents, or faces only a small state space, then it may be feasible to represent utility functions exactly (i.e. use $\hat{\mathbf{u}}^{(m)} = \mathbf{u}^{(m)}$), even for $\lambda > 0$. But if the agent population or state space is large, then the optimal representation in terms of cost-adjusted utility (Eq. \ref{eq:cost-adjusted-utility}) will likely be an approximation that discards some information for the sake of lower entropy (see Appendix \ref{appx:entropy}). A simple way to obtain such an approximation is through state aggregation---i.e. group all states within a given-sized `patch' of the grid under a single value \citep{SuttonBarto, Abel2019StateAA}. Fig. \ref{fig:2d-bandit-results} shows how the cost-balanced utility of aggregated value function representations under various values of the tradeoff parameter $\lambda$. As expected, we observe that increasing the amount of aggregation leads to a decrease in both average return and representation cost, with the relative advantage or disadvantage determined by the tradeoff parameter.

\section*{Experiment 2: representing social groups}
\begin{figure}
    \centering
    \includegraphics[width=12.5cm]{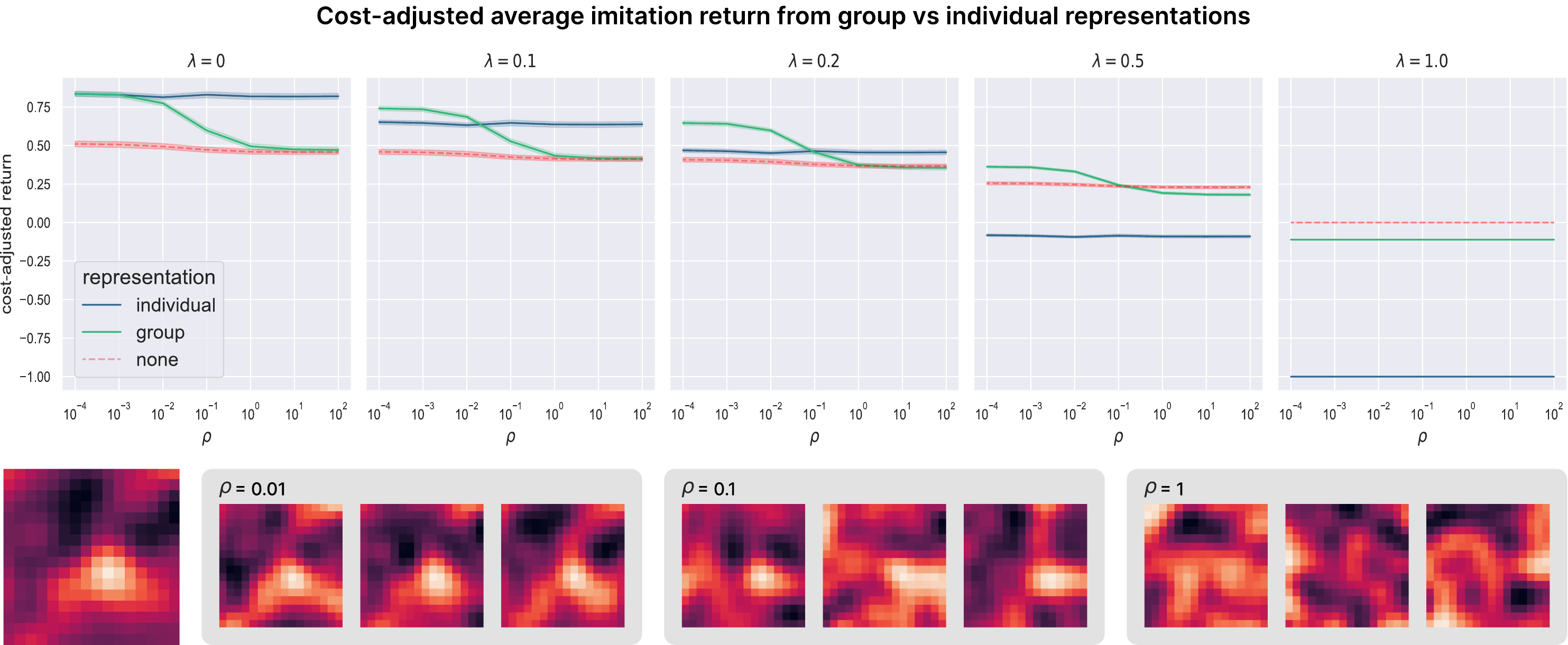}
    \caption{Top: cost-adjusted return for group-only and individual representations (with indiscriminate baseline) for different values of the group variance ratio $\rho$ and the tradeoff parameter $\lambda$. As before, both return and cost (entropy) are normalised to lie in [0,1]. Bottom: illustration of example utility functions sampled for a 3-member group for three different values of the variance ratio $\rho$.}
    \label{fig:2d-bandit-results-groups}
\end{figure}

So we have seen how naive compression of utility function information can help a social learner navigate this tradeoff. But rather than merely taking individual representations and making them coarser, a more \emph{humanlike} approach might involve using information about the agent population \emph{as a whole}, and how individual agents within the population relate to one another. For instance, we often represent people that we're unfamiliar with in terms of their membership of certain social \emph{groups} or \emph{categories}, using our knowledge of those groups to make inferences about the unknown properties of their members \citep{Rhodes2019TheDO, Liberman2017TheOO}. To see how this can produce cheap agent representations, imagine that we have a known fixed number of groups $K$, and a population of $M$ agents that are partitioned such that each agent $m$ belongs to only one group (given by $\mathbf{z}_m$). Continuing our example of selective social learning, we will say that each group $k$ is associated with a corresponding distribution over utility functions, $p(\mathbf{u}^{(m)} |\mathbf{z}_m = k)$, with mean value $\mathbf{\mu}^{(k)}$. As an extreme, if we treat all members of group $k$ as having $\mathbf{u} = \mathbf{\mu}^{(k)}$ (i.e. representing zero intragroup variance), then we need only represent each agent with a single scalar value $\mathbf{z}_m$. Including the group means, this would yield an overall representation cost
\begin{equation}
    C(\chi_\text{grp}) = KH[\mathbf{\mu}^{(k)}] + MH[\mathbf{z}_m]
\end{equation}
Intuitively, the extent to which this representation is useful will be primarily determined by the ratio between inter- and intragroup variation. When groups are very different, and agents within each group are very similar, then knowing an agent's group assignment tells you a lot about their individual properties; when groups are similar or agents vary a lot within each group, then group assignments are less informative. We can test this using the same task setup as in the previous experiment. We first sample a number of group mean value functions $\mathbf{\mu}^{(k)}$ from the original GP prior; for each group we then sample a number of agents from a multivariate Gaussian distribution with mean $\mathbf{\mu}^{(k)}$ and covariance $\Sigma = \rho\Sigma_\text{GP}$ where $\rho > 0$ is a scalar and $\Sigma_\text{GP}$ is the covariance of the GP prior. Fig. \ref{fig:2d-bandit-results-groups} compares the cost-adjusted utility of the groups-only and individuals-only representation strategies for different values of $\rho$ and $\lambda$. We can see that, below a certain $\lambda$ threshold, individual representations are always as good or better than group representations; above a certain $\lambda$ threshold, the reverse is true. Otherwise, as predicted, the optimal representation depends on the group variance ratio, with group-only representations becoming less useful as $\rho$ increases. 

\section*{Discussion}
In this paper, we have briefly laid out and motivated the problem of cost-constrained social representation; that is, representing other agents in such a way that optimally trades off between downstream utility and information cost. Using selective social learning as an example task, we explored two potential approaches to this problem: compressing individual representations, and representing agents in terms of group identity. While we hope the work we present here is of some value, it is of course very much a preliminary step---in future work we hope to consider how agents can \emph{learn} good social representations by optimising directly for this tradeoff, as well as broadening our analysis to encompass additional example tasks and agent features. As hinted at earlier, we also hope to explore a more detailed conception of representation cost that considers both acquisition and storage of information. 

\bibliographystyle{abbrvnat}
\bibliography{references} 

\appendix
\section{Representation entropy}\label{appx:entropy}
Here we provide some additional details on our use of representation entropy as a measure of cognitive cost. For a comprehensive introduction to entropy (and other related information-theoretic quantities) we direct the reader to \citet{CoverThomasElements}. For a discrete random variable $X \in \mathcal{X}$ with probability mass function $p(x) := \text{Pr}(X = x)$, the entropy of $X$ is given by
\begin{equation}\label{eq:discrete-entropy}
    H_b(X) = -\sum_{x\in\mathcal{X}}p(x)\log_b(p(x))
\end{equation}
When $b = 2$ (as it typically is), the entropy has units of bits. For a continuous random variable $Y \in \mathcal{Y}$ with probability density function $p(y)$, the differential entropy of $Y$ is given by 
\begin{equation}\label{eq:continuous-entropy}
    H_b(Y) = -\int_{y\in\mathcal{Y}}p(y)\log_b(p(y))
\end{equation}
If $Y$ follows a multivariate Gaussian distribution with covariance $\Sigma$, then $H(Y)$ is computed as
\begin{equation}\label{eq:multivariate-gaussian-entropy}
    H_2(Y) = \frac{1}{2}\log_2|\Sigma| + \frac{n}{2}\big(\log_2(2\pi e)\big) 
\end{equation}
where $n$ is the dimensionality of $Y$ \citep{RasmussenWilliamsGaussian}. This allows us to compute the entropy of our agent value functions $\mathbf{v}$, since each value function is in effect a continuous random variable distributed according to a multivariate Gaussian with known $\Sigma$. The total entropy of a representation $\chi_\text{ind}$ that consists of the value functions of $M$ agents over $n$ states is thus given by
\begin{equation}\label{eq:chi-individual-entropy}
    H_2[\chi_\text{ind}] = \frac{M}{2}\big[\log_2|\Sigma| + n\big(\log_2(2\pi e)\big) \big]
\end{equation}

To compute the entropy of a state-aggregated value function estimate $\hat{\mathbf{v}}$, we can treat $\hat{\mathbf{v}}$ as a new continuous RV with a lower-dimensional multivariate Gaussian distribution whose covariance $\Sigma_\text{agg}$ is determined entirely by $\Sigma$ and the level of state aggregation. Determining $\Sigma_\text{agg}$ is then sufficient to compute $H(\hat{\mathbf{v}})$. For the group-only representation cost, we need simply to compute the entropy of the scalar discrete random variable $\mathbf{z}$ (the agents' group assignments). With a uniform distribution over $K$ groups, this is simply given (following Eq. \ref{eq:discrete-entropy}) by
\begin{equation}\label{eq:group-entropy}
    H_2(\mathbf{z}) = -\sum_{k=1}^K \frac{1}{K}\log_2\big(\frac{1}{K}\big) = \log_2(K)
\end{equation}
The total entropy of a groups-only representation $\chi_\text{grp}$ is given by combining this with the total entropy of the group mean value functions:
\begin{equation}\label{eq:chi-group-entropy}
    H_2[\chi_\text{grp}] = \frac{K}{2}\big[\log_2|\Sigma| + n\big(\log_2(2\pi e)\big) \big] + M\log_2(K)
\end{equation}

\end{document}